# Concise Fuzzy System Modeling Integrating Soft Subspace Clustering and Sparse Learning

Peng Xu, Zhaohong Deng, *Senior Member, IEEE*, Chen Cui, Te Zhang, Kup-Sze Choi, *Member, IEEE*, Suhang Gu, Jun Wang, Shitong Wang

*Abstract*—The superior interpretability and uncertainty modeling ability of Takagi-Sugeno-Kang fuzzy system (TSK FS) make it possible to describe complex nonlinear systems intuitively and efficiently. However, classical TSK FS usually adopts the whole feature space of the data for model construction, which can result in lengthy rules for high-dimensional data and lead to degeneration in interpretability. Furthermore, for highly nonlinear modeling task, it is usually necessary to use a large number of rules which further weakens the clarity and interpretability of TSK FS. To address these issues, a concise zero-order TSK FS construction method, called ESSC-SL-CTSK-FS, is proposed in this paper by integrating the techniques of enhanced soft subspace clustering (ESSC) and sparse learning (SL). In this method, ESSC is used to generate the antecedents and various sparse subspace for different fuzzy rules, whereas SL is used to optimize the consequent parameters of the fuzzy rules, based on which the number of fuzzy rules can be effectively reduced. Finally, the proposed ESSC-SL-CTSK-FS method is used to construct concise zero-order TSK FS that can explain the scenes in high-dimensional data modeling more clearly and easily. Experiments are conducted on various real-world datasets to confirm the advantages.

## I. Introduction`

Fuzzy system (FS) is a kind of rule-based systems which use fuzzy logic and fuzzy inference to realize knowledge representation and uncertain inference. The core part of FS is the knowledge base which is composed of IF-THEN fuzzy rules. FS has better ability in handling uncertainty than other rule-based systems. It can transform vague human language into fuzzy rules and simulate the uncertain inference. In various existing FSs, TSK FS [1-4] is the one that is most commonly used because of two major advantages: the intrinsic interpretability of its rule-based form and the data-driven learning ability.

Most of the recent researches focus on the improvement of the learning ability of TSK FS. However, as model complexity increases, the interpretability of TSK FS decreases despite increase in learnability [5-7]. It is therefore imperative to revisit the interpretability of TSK FS, which is among the emerging topics on the interpretabilities of machine learning methods [8]. There are mainly two classes of interpretable methods in the field of machine learning [9]. The first class is intrinsically interpretable models, where linear models, decision trees and rule-based systems are common models of this class. The second class is post-hoc interpretability methods that usually apply interpretable methods to extract information from the learned black-box models. Common models of the second class include visualizations [10], explanations by examples [11] and model-agnostic methods [12]. The TSK FS concerned in this paper belongs to the first class. On the other hand, it is noteworthy to highlight that TSK-FS has stronger learning abilities than other types of FSs and is more capable of dealing with uncertain information than other non-fuzzy rule-based systems [13].

As pointed out in [9], given the limited capacity of human cognition, neither rule-based systems nor decision trees are interpretable with increasing complexity of the models. While TSK FS is interpretable for its rule-based form, the interpretability can be reduced to a great extent as model complexity increases. There are two major factors that could increase the model complexity of TSK FS: 1) *High-dimensional features*: classical data-driven TSK FSs usually use all input features to generate fuzzy rules. In fact, only partial features are useful for certain fuzzy rules. Besides, fuzzy rules generated using all the features would be very long, and the corresponding linguistic descriptions are tedious, which reduces the interpretability of the fuzzy models. 2) *Excessive rules*: To achieve good performance, many rules are usually required to construct TSK FS, which inevitably increase the model complexity. In fact, some rules may be redundant and can be removed without losing modeling performance. Thus, concise rule base is necessary to improve the interpretability.

To deal with the abovementioned issues, a more interpretable TSK FS construction method is proposed in the paper. The method first introduces an enhanced soft subspace clustering (ESSC) [14] method to perform feature reduction for each fuzzy rule, where ESSC can select different feature subsets for different fuzzy rules. Then, sparse learning (SL) [15] technique is used to optimize the consequent parameters of fuzzy rules to perform rule reduction. Finally, a more concise TSK FS is obtained with the number of features and fuzzy rules reduced simultaneously. The proposed is thus called enhanced soft subspace clustering and sparse learning based concise zero-order TSK FS,

This work was supported in part by the National Key Research Program of China under Grant No. 2016YFB0800803, the NSFC under Grant No. 61772239, the Jiangsu Province Outstanding Youth Fund under Grant No. BK20140001, by the National First-Class Discipline Program of Light Industry Technology and Engineering under Grant LITE2018-02, and the Hong Kong Research Grants Council under Grant No. PolyU 152040/16E. (Corresponding author: Zhaohong Deng)

Peng Xu, Zhaohong Deng, Chen Cui, Te Zhang, Suhang Gu, Jun Wang, Shitong Wang are with the School of Digital Media, Jiangnan University and Jiangsu Key Laboratory of Digital Design and Software Technology, Wuxi 214122, China (e-mail: 6171610015@stu.jiangnan.edu.cn; dengzhaohong@jiangnan.edu.cn; 569398535@qq.com; ztcsrookie@163.com; gusuhang09@163.com; wangjun_sytu@sina.com; wxwangst@aliyun.com).

K.S. Choi is with The Centre for Smart Health, School of Nursing, the Hong Kong Polytechnic University (e-mail: thomasks.choi@polyu.edu.hk)

abbreviated as ESSC-SL-CTSK-FS. It has the following advantages:

1) The TSK FS is trained using ESSC and SL simultaneously so that the fuzzy rules are more linguistically descriptive and the rule base is more compact, thereby increasing the interpretability of the model.

2) The fuzzy model constructed possesses a more human-like inference mechanism that different feature subsets are used in different rules by the ESSC for fuzzy inference, which is analogous to making a decision based on the views of different experts.

3) The ESSC clustering algorithm avoids the use of noisy features in the original feature space by unsupervised learning. Therefore, the proposed method is more robust in handling high-dimensional data where noisy features may be embedded in the original feature space.

The rest of this paper is organized as follows. In section II, the related work on the interpretability of fuzzy systems is briefly reviewed. Zero-order TSK FS, ESSC and SL are then introduced in Section III. In section IV, the proposed method is discussed in detail. The experimental studies are reported in section V. Finally, conclusions and future work are given in Section VI.

## II. RELATED WORK

### A. Interpretability

As discussed in Section I, intrinsically interpretable models are an important class of interpretable methods. The representative models of this class are linear models, decision trees and rule-based systems [16]. *Transparency* is a key property of these models, which can be considered from three levels [9], i.e., *simulatability* at the level of the entire model, *decomposability* at the level of individual components, and *algorithmic transparency* at the level of the training algorithm. But the transparency of these models can be damaged by various factors. The interpretability of linear regression models can be weakened by the high-dimensional features and Lasso is often used to introduce sparsity into the model for better interpretability [15]. The decision trees are interpretable due to their induced decision sets [16], but the interpretability of renowned tree ensembles are limited, such as random forest and boosted trees [17]. Both researches [18] and [19] propose methods to make tree ensembles more interpretable. Rule-based systems are probably the most interpretable models because their IF-THEN structure semantically resembles natural language and the way human think [20]. There are also many researches to construct the more interpretable rule-based systems [21, 22].

Besides investigations based on the intrinsically interpretable models as discussed above, another line of research on interpretability is post-hoc explanations of black-box models (e.g., deep neural networks). Here, neural network visualization is an important research direction [10]. Previous researches in [23] and [24] both visualize the individual units of neural networks to understand their representations. In [25] and [26], methods are proposed to disentangle the representations and quantify the interpretability of neural networks. Some researches attempt to interpret the models at the level of examples. Counterfactual explanations are used to describe how the changes to the features of an example can change the prediction of a black-box model to a predefined output [27]. Adversarial examples are another research hotspot of example-based interpretability, which is not to interpret a model like counterfactual examples but to deceive it [28, 29]. Many other model-agnostic methods are also proposed to improve the interpretability of machine learning algorithms, such as local surrogate models [30] and influence functions [31].

### B. Fuzzy Systems

For the TSK FS belonging to the intrinsically interpretable models, its transparency is analyzed according to [9] as follows. The simulatability is clearly evident from the rule-based human-like inference mechanism [32]. The decomposability is evident from the characteristic that all the components of the TSK FS can provide intuitive explanations. The training algorithm of it can converge to a unique solution [1], which exhibits the property of algorithmic transparency. While with the increase of the number of rules in the rule base and the length of rules, the transparency of TSK FS is decreased.

Researches have been conducted to improve the interpretability of TSK FS with concise rule base. In [33], Type-2 hierarchical fuzzy system (T2HFS) is proposed for handling high-dimensional data, where principle component analysis is used for feature extraction. However, the interpretability is weakened since the physical meaning of the original features is destroyed. In [34], genetic algorithm and integer programming are integrated to propose a fuzzy rule classifier algorithm to tackle the precision and rule reduction problems in the classification of high-dimensional data. The method does not consider the damage in interpretability caused by high dimensionality. In [32, 35, 36], subsets of features have been used in different rules, but the selection of features for the subsets is often conducted randomly which decreases the effectiveness of the constructed model. In [32], a TSK FS called ETSK-FS is proposed to construct more concise fuzzy model by using the extracted feature subsets for different fuzzy rules. However, the removal of redundant rules is not addressed. In summary, it is necessary to develop adaptive methods to improve the performance of TSK FS, both the prediction accuracy and the model conciseness.

## III. BACKGROUNDS

### A. Zero-order TSK-FS

TSK fuzzy logic system (TSK FS) [37] is a classical fuzzy inference model that has been widely applied because of its flexibility and performance. In this paper, the commonly used zero-order TSK FS is investigated due to its simplicity. For a zero-order TSK FS, the fuzzy rules are defined as follows [38].

$$\begin{aligned}&\text{IF } x_1 \text{ is } A_1^k \wedge x_2 \text{ is } A_2^k \wedge \cdots \wedge x_d \text{ is } A_d^k, \\ &\text{THEN } y^k(\mathbf{x}) = p^k, \qquad k = 1, \cdots, K.\end{aligned} \quad (1)$$

The $k$th rule in the fuzzy rule base is given in (1), where $A_i^k$ is the fuzzy subset associated with the $i$th feature, $p^k$ is the consequent parameter, $y^k$ is the output of this rule, $k = 1, 2, ..., K$, and $K$ is the number of fuzzy rules in the rule base. For an input vector $\mathbf{x} = (x_1, x_2, ..., x_d)^T$, when multiplication is adopted for conjunction and implication, addition for combination, and the center of gravity for defuzzification, the output of the zero-order TSK FS, i.e., $f(\mathbf{x})$, can be expressed as

$$f(\mathbf{x}) = \sum_{k=1}^{K} \frac{u^k(\mathbf{x})}{\sum_{k'=1}^{K} u^{k'}(\mathbf{x})} y^k(\mathbf{x}) = \sum_{k=1}^{K} \tilde{u}^k(\mathbf{x}) y^k(\mathbf{x}). \quad (2)$$

In (2), $u^k(\mathbf{x})$ and $\tilde{u}^k(\mathbf{x})$ are commonly called the firing strength and the normalized firing strength respectively, which are given by

$$u^k(\mathbf{x}) = \prod_{i=1}^{d} u_{A_i^k}(x_i) \text{ and} \quad (3)$$

$$\tilde{u}^k(\mathbf{x}) = u^k(\mathbf{x}) / \sum_{k=1}^{K} u^k(\mathbf{x}), \quad (4)$$

where $u_{A_i^k}(\cdot)$ is the membership function of fuzzy subset $A_i^k$ associated with the $i$th feature in the $k$th rule. Gaussian function is commonly used as the membership function [38], i.e.,

$$u_{A_i^k}(x_i) = \exp(\frac{-(x_i - v_i^k)^2}{2\sigma_i^k}). \quad (5)$$

The parameters $v_i^k$ and $\sigma_i^k$ in (5) can be estimated with different strategies, such as clustering techniques. If fuzzy c-mean (FCM) is used, they can be obtained by

$$v_i^k = \sum_{j=1}^{N} u_{jk} x_{ji} / \sum_{j=1}^{N} u_{jk} \text{ and} \quad (6)$$

$$\sigma_i^k = h \sum_{j=1}^{N} u_{jk} (x_{ji} - v_i^k)^2 / \sum_{j=1}^{N} u_{jk}, \quad (7)$$

where $u_{jk}$ represents the membership of the input vector $\mathbf{x}_j = (x_{j1}, x_{j2}, ..., x_{jd})^T$ of the $k$th cluster and $h$ is a manually adjustable scale parameter [39, 40].

Once the antecedent parameters in the zero-order TSK FS are determined, the output can be expressed as a linear model in a new feature space as follows.

$$y = f(\mathbf{x}) = \mathbf{p}_g^T \mathbf{x}_g \quad (8)$$

$$\mathbf{x}_g = [\tilde{u}^1(\mathbf{x}), \tilde{u}^2(\mathbf{x}), ..., \tilde{u}^K(\mathbf{x})]^T \quad (9)$$

$$\mathbf{p}_g = [p^1, p^2, ..., p^K]^T \quad (10)$$

### B. Soft Subspace Clustering

Clustering has been widely applied for fuzzy modeling. It is effective for partitioning the training data to generate the appropriate space partition. Based on the partition results, the corresponding parameters of the antecedents and/or consequents in the fuzzy rules can be estimated. For example, a fuzzy rule can be generated based on a group in the clustering results. Traditional clustering algorithms, such as K-means [41] and fuzzy c-means FCM [42], have been used in many fuzzy modeling methods to determine the space partition and generate fuzzy rules. However, these clustering methods usually generate fuzzy rules using the same feature space, which is not reasonable for many practical applications. For a rule generated based on the knowledge of an expert, the rule only contains a certain feature subset that is associated with the view of the expert. When multiple experts are involved, it is therefore more appropriate for the different rules to be associated with different feature subsets. To this end, adaptive clustering techniques are needed. One example is the soft subspace clustering (SSC) technique [14] which is originally proposed to effectively cluster high-dimensional data. A distinctive characteristic of SSC is that it can identify different groups and determine the importance of the features in each group simultaneously. The optimization objective function of different SSC algorithms can be expressed with the generalized form:

$$J(\mathbf{U}, \mathbf{V}, \mathbf{W}) = f_1(\mathbf{U}, \mathbf{V}, \mathbf{W}) + \Delta, \quad (11)$$

where $\mathbf{U}$ is a partition matrix of data, $\mathbf{V}$ is a matrix containing $K$ center vectors, $\mathbf{W}$ is a weight matrix that characterizes the importance of features in different clusters. The objective function consists two terms, the loss term $f_1$ and the regularization term $\Delta$.

Theoretically, the more compact a cluster is in some dimensions of the feature space, the larger the corresponding weights. Thus, important feature subsets can be identified based on the clustering results of the SSC algorithms, which can be adopted to generate the fuzzy rules in different feature subspace.

### C. Lasso Sparse Learning

Sparse learning has received increasing attention in recent years for intelligent modeling. Lasso is a classical sparse learning method for linear regression model [43]. Its aim is to obtain a sparse vector as the solution to a linear model such that the final decision function of the linear model is concise. Given a training dataset $\{\mathbf{x}_i, y_i\}, i = 1, 2, ..., n$ for a linear model $f(\mathbf{x}) = \mathbf{w}^T \mathbf{x}$, the optimization objective of the Lasso algorithm is given by

$$\min_{\mathbf{w}} \sum_{i=1}^{n} (y_i - \mathbf{w}^T \mathbf{x}_i)^2 + \lambda \| \mathbf{w} \|_1, \quad (12)$$

where $\| \mathbf{w} \|_1$ is the $\ell_1$ norm of $\mathbf{w}$ and $\lambda (> 0)$ is the regularization parameter. The $\ell_1$ norm of $\mathbf{w}$ is introduced into the learning criterion to effectively reduce over-fitting which may occur in dealing with small high-dimensional dataset. Besides, it is apt to obtain a sparse solution for the linear model.

## IV. CONCISE TSK FS CONSTRUCTION USING ESSC AND SL

In this paper, a concise TSK FS is constructed using SSC and SL techniques. The fuzzy rules of the concise TSK FS are designed as follows,

$$\begin{aligned} &\text{IF } x_1^k \text{ is } A_1^k \wedge x_2^k \text{ is } A_2^k \wedge \cdots \wedge x_{m_k}^k \text{ is } A_{m_k}^k, \\ &\text{THEN } y^k = f^k(\mathbf{x}^k), \qquad k = 1, \cdots, K. \end{aligned} \quad (13)$$

where $K$ is the number of rules, $\mathbf{x}^k = (x_1^k, x_2^k, ..., x_{m_k}^k) \in R^{m_k}$ is a vector containing $m_k$ features that are extracted from the full $d$ features of the original input vector $\mathbf{x}$, $A_i^k$ is the fuzzy subset associated with feature $x_i^k$ in the $k$th fuzzy rule, $f^k(\mathbf{x}^k)$ is the consequent output, which is a constant in the adopted zero-order TSK FS, i.e., $f^k(\mathbf{x}^k) = p^k$.

### A. Antecedent Parameter Estimation using ESSC

Many SSC algorithms have been proposed to cluster high-dimensional data and to find the important subspace for each group [44-46]. Unlike most existing SSC algorithms that only focus on the within-cluster compactness, ESSC exhibits distinctive advantage that it considers not only the within-cluster compactness but also the between-cluster separation simultaneously [14]. Given the training data $\mathbf{X}=[\mathbf{x}_1,\mathbf{x}_2,\cdots,\mathbf{x}_N]$ and corresponding labels $\mathbf{y}=[y_1,y_2,\cdots y_N]^T$, where $\mathbf{x}_i \in R^d$ and $y_i \in R$, the loss function of the ESSC algorithm is:

$$J_{\text{ESSC}}(\mathbf{U},\mathbf{V},\mathbf{W}) = \sum_{i=1}^{K}\sum_{j=1}^{N} u_{ij}^m \sum_{k=1}^{d} w_{ik}(x_{jk}-v_{ik})^2 + \varepsilon \sum_{i=1}^{K}\sum_{k=1}^{d} w_{ik}\ln w_{ik}$$
$$- \eta \sum_{i=1}^{K}(\sum_{j=1}^{N} u_{ij})\sum_{k=1}^{d} w_{ik}(v_{ik}-v_{0k})^2 \quad , \quad (14)$$

$$\text{s.t. } 0 \leq u_{ij} \leq 1, \ \sum_{i=1}^{K} u_{ij}=1, \ 0 < \sum_{j=1}^{N} u_{ij} < N, \ 0 \leq w_{ik}, \ \sum_{k=1}^{D} w_{ik}=1$$

where $\mathbf{V}=[\mathbf{v}_1,...,\mathbf{v}_K]_{d\times K}$ and $\mathbf{W}=[\mathbf{w}_1,...,\mathbf{w}_K]_{d\times K}$ are the matrix of clustering center and the weighting matrix respectively, and $\mathbf{U}=[\mathbf{u}_1,...,\mathbf{u}_N]_{K\times N}$ is the fuzzy partition matrix. $K, N, d$ are the number of clusters, the number of samples and the number of features respectively. The objective function of ESSC consists of three terms which correspond to the weighted within-cluster compactness, regularization of feature weights, and the weighted between-cluster separation respectively. The first and second terms directly inherit from the objective function of the classical SSC algorithm FSC [47]. The parameters $\varepsilon, \eta$ are used to balance the influence of the different terms.

As a state-of-the-art SSC algorithm, ESSC is very effective for partitioning high-dimensional data, where the corresponding distribution of the importance of the different groups can be also be obtained. In this study, ESSC is used to generate the antecedents of the zero-order TSK FS so that more elastic fuzzy rules with different feature subsets can be constructed. The procedure of the antecedent generation is described as follows.

For a given training dataset, the ESSC method divides the training examples into $K$ groups that correspond to the $K$ fuzzy rules; the weight vector $\mathbf{w}_k=[w_{k1},\cdots,w_{kd}]$ corresponds to the contribution of all the features to the $k$th group. Given the threshold $\beta \in (0,1)$, for the $k$th rule, the features with the weights $w_{kl} > \beta$ are selected. When the feature subsets for each fuzzy rule are determined, the antecedent parameters, i.e., the parameters $v_i^k$ and $\sigma_i^k$ in the Gaussian membership function, can be estimated using (6) and (7) as described in Section II-A.

B. *Consequent Parameter Estimation and Rule Reduction using SL*

After generating the antecedents, the consequent parameters of the fuzzy rules of the zero-order TSK FS can be estimated based on linear model optimization techniques. According to the descriptions in Section II-A, we construct the following objective function for optimization:

$$\min_{\mathbf{p}_g} \sum_{i=1}^{N}(y_i - \mathbf{p}_g^T \mathbf{x}_{i,g})^2 \ , \quad (15)$$

where $\mathbf{x}_{i,g} \in R^K$ is the input vector in the new feature space, which is mapped from the input vector $\mathbf{x}_i \in R^d$ in the original feature space through (9) with the selected features obtained by ESSC, and $\mathbf{p}_g$ is the linear model parameters in the new feature space, which is composed of the consequent parameters of all the fuzzy rules. Each element in $\mathbf{p}_g$ is associated with a fuzzy rule.

For zero-order TSK FS, the number of fuzzy rules may be large and the resulting model would become more complicated and less interpretable. However, a concise model for TSK FS is generally expected in practical applications. To this end, Lasso algorithm [43] is introduced for both consequent parameters estimation and fuzzy rules reduction. The objective function is given by

$$\min_{\mathbf{p}_g} \frac{1}{2}\|\mathbf{X}_g \mathbf{p}_g - \mathbf{y}\|^2 + \frac{\lambda}{2}\|\mathbf{p}_g\|_1 \ . \quad (16)$$

In (16), $\mathbf{X}_g=[\mathbf{x}_{1,g},\mathbf{x}_{2,g},\cdots,\mathbf{x}_{N,g}]^T$ is obtained by concatenating the data from all the examples in the new feature space. $\|\mathbf{p}_g\|_1$ denotes the $\ell_1$ norm of $\mathbf{p}_g$, which is opt to produce a sparse solution vector, i.e., some elements of $\mathbf{p}_g$ is reduced to zero; $\lambda$ is a regularization parameter to control the influence of $\|\mathbf{p}_g\|_1$. When $\lambda$ is set appropriately, the solution $\mathbf{p}_g^*$ only contains a small number of nonzero elements, i.e., the consequent parameters of many rules in the zero-order TSK FS are zero. Furthermore, when the consequent parameter of a fuzzy rule is equal to zero, the rule can be removed. Note that the objective function in (16) is non-smooth due to the non-smoothness of the $\ell_1$ norm regularization term. The accelerated proximal gradient descent method is used to solve this problem, where the solution $\mathbf{p}_g=[p^1,p^2,...,p^K]^T$ can be learned with the following update rules [48]

$$p_{g_s} = \begin{cases} p_{g_s} - \gamma, & p_{g_s} > \gamma \\ 0, & |p_{g_s}| \leq \gamma \\ p_{g_s} + \gamma, & p_{g_s} < -\gamma \end{cases}, \quad (17)$$
$$s=1,2,\cdots,K$$

where $\text{sgn}(.)$ is the sign function and $\gamma$ is a positive constant. Here, $\gamma$ is obtained by $\gamma = \lambda/L$ and $L$ is the Lipschitz constant which can be calculated using the approach in [48]. The optimization procedure in (17) is iterative and the initial $\mathbf{p}_g^*$ can be obtained as an analytic solution to $\min_{\mathbf{p}_g}\|\mathbf{X}_g \mathbf{p}_g - \mathbf{y}\|^2$. The iteration is repeated until $\mathbf{p}_g$ converges.

C. *The ESSC-SL-CTSK-FS Algorithm*

The ESSC-SL-CTSK-FS algorithm proposed in this paper to improve the interpretability of FS is presented in Table I. First, ESSC is used to select important subspace for the antecedents. Lasso sparse learning is then used to learn the consequent parameters and reduce the number of rules.

D. *Multi-Class Classification*

TSK FSs have been widely applied for regression and classification. In most cases, fuzzy regression models can be constructed easily, whereas fuzzy classification problems are solved using fuzzy regression models with multiple outputs. Given a multi-class data set $D=\{\mathbf{x}_j,y_j\}, y_i \in \{1,...,m\}$, a regression dataset with multiple outputs $\{\mathbf{x}_i, \tilde{\mathbf{y}}_i\}$ can be constructed, where $y_i \in R^m$ is the output vector for the $i$th sample. If the label of the $i$th sample is $p \ (1 \leq p \leq m)$, $y_i$ is encoded as $\tilde{y}_i=[0,...,1,0,...,0]^T$, where the $p$th element of $y_i$ is 1 and the remaining elements are 0. In this way, the original classification problem is transformed into $m$ regression problems and $m$ fuzzy regression models can thus be built. For a test sample, the outputs of the $m$ regression models can be encoded as $\tilde{y}_i^{model} = [y_{i,1}^{model},...,y_{i,m}^{model}]^T$ which is assigned to the class corresponding to the largest output element in $y_i^{model}$.

### E. Complexity Analysis

#### 1) Model Complexity

The model complexity is evaluated by the number of model parameters in the final model. In this paper, the trained zero-order TSK FS contains antecedent parameters $v_i^k$, $\sigma_i^k$ in the fuzzy membership functions, and the consequent parameter $p^k$. The antecedent parameters and the non-zero consequent parameters determine the complexity of the final fuzzy model. For traditional zero-order TSK, the number of parameters in the final model is $(2d+1)K$, where $d$ is the number of features, and $K$ is the number of fuzzy rules. For the proposed ESSC-SL-CTSK-FS, the trained zero-order TSK FS has more elastic rules, i.e., different feature subsets are used for the antecedents in different rules. Note that the consequent parameters of some rules may be zero after Lasso sparse learning. Hence, the number of final rules is smaller than $K$. Denote the number of final rules as $K'$ and let $m_k$ be the number of selected features for the $k$th rule whose consequent parameter is not equal to zero, the maximum model complexity is $\sum_{k=1}^{K'}(2m_k+1)$.

#### 2) Time Complexity

The proposed ESSC-SL-CTSK-FS algorithm includes two major steps: the acquisition of antecedent parameters of the fuzzy rules using ESSC, and the learning of the consequent parameters using Lasso algorithm. In the first step, the time complexity of ESSC is $O(TNKd)$ where $T$, $N$, $K$ and $d$ are the number of iterations, data, clusters and features respectively. The second step is essentially a classical Lasso problem that can be solved by many existing methods [15, 49, 50]. In this paper, the accelerated proximal gradient method [48] is adopted and the time complexity is detailed in [51, 52]. Conclusively, the time complexity of ESSC-SL-CTSK-FS is $O(TNKd + N)$.

## V. Experiments

### A. Experimental Setup

To evaluate the effectiveness, the proposed ESSC-SL-CTSK-FS is compared with several classical fuzzy models, including TSK fuzzy classification (TSK-FC) [53], L2-norm penalty-based TSK FS (L2-TSK-FS) [1], TSK FS based on IQP optimization (TSK-IQP) [1], TSK FS based on LSSLI optimization (TSK-LSSLI) [1], and PCA feature extraction based L2-TSK-FS (TSK-FS-PCA) [1].

For all the methods under comparison, the hyper-parameters are optimized using five-folds cross-validation and grid search strategy. Since all the methods are FS based, their parameters are almost the same. The following arrangements are made for optimal parameter setting. The number of fuzzy rules is set to 30. The scale parameter in the Gaussian function is set using the search grid $\{0.01,0.1,1,10,100\}$. The regularization parameter is set using the search grid $\{2^{-10}, 2^{-9},\cdots,2^0,\cdots,2^9, 2^{10}\}$. For the proposed method, the threshold parameter for feature selection is set using the search grid $\{0.1, 0.15, 0.2, 0.25, 0.3\}$ with a step size of 0.05; the regularization parameters $\varepsilon$ and $\eta$ in (10) are set

TABLE I
FLOWCHART OF THE ESSC-SL-CTSK-FS ALGORITHM

**Algorithm**: ESSC-SL-CTSK-FS

**Input**: the number of fuzzy rules $K$; the adjustment parameter $h$ in the Gaussian membership function; the weight threshold $\beta$ for feature selection; the regularization parameter $\eta$ in (14); and the training data $D_{tr} = \{x_i, y_i\}$.

**Output**: antecedent $v_i^k$ and $\sigma_i^k$, consequent parameters $\mathbf{p}_g$.

**Procedure ESSC-SL-CTSK-FS**:

**Stage 1: Generation of concise antecedents using ESSC**

  **Step 1**: Implement ESSC on the input dataset $\{\mathbf{x}_i\}$. Divide $\{\mathbf{x}_i\}$ into $K$ clusters and obtain the partition matrix $\mathbf{U}$. Set the cluster center matrix $\mathbf{V}$ and the feature weight matrix $\mathbf{W}$.

  **Step 2**: Match each cluster to a fuzzy rule. Determine the importance of the features for each rule using $\mathbf{W}$ and $\beta$.

  **Step 3**: Estimate the parameters of the fuzzy membership functions with (6) and (7).

**Stage 2: Consequent parameter learning and rule reduction using LASSO**

  **Step 4**: Set the regularization parameter $\lambda$ in (16).

  **Step 5**: Solve the optimal consequent parameters using (17).

**Stage 3: TSK FS construction**

  **Step 6**: The final TSK FS is constructed using the antecedent and consequent parameters obtained in Stage 1 and Stage 2.

using search grids $\{0.01,0.1,1,10,100\}$ and $\{0.01,0.05,0.1,0.3,0.5\}$ respectively. The regularization parameter $\lambda$ in (16) is set using the search grid $\{0.1, 0.2, \cdots, 0.8, 0.9\}$.

### B. Datasets

Ten real-world medical datasets from the UCI Repository are adopted for performance comparison. In the experiments, all the features of the samples are normalized to the interval [0,1]. Table II gives the details of the datasets, including the name of the datasets, the number of samples and features in the dataset, the number of classes and the number of samples in each class.

### C. Evaluation Index

The proposed ESSC-SL-CTSK-FS is evaluated from the perspectives of classification performance and model complexity.

#### 1) Classification Performance

Six metrics are adopted to evaluate the classification performance with five-fold cross-validation, i.e., accuracy, precision (P), recall (R), F-measure, rand index (RI) and Jaccard measure. Accuracy is defined based on the four elements of the confusion matrix, i.e., true positive (TP), false positive (FP), true negative (TN) and false negative (FN), and is given as follows.

$$Accuracy = (TP+TN)/(TP+FP+FN+TN) \qquad (18)$$

Since accuracy cannot reflect the situation of imbalanced data [54], F-measure is also adopted. As shown in (19), F-measure combines P and R to provide more insight into the functionality of a classifier. The macro-averaged P, R and F-measure are adopted in the experiments for multi-class tasks [55].

$$P = TP/(TP+FP) \qquad (19a)$$

$$R = TP/(TP+FN) \qquad (19b)$$

$$F\text{-}Measure = (2 \times P \times R)/(P+R) \qquad (19c)$$

TABLE II
TEN REAL-WORLD MEDICAL DATASETS

| Index | Datasets | Samples | Features | Classes (+/-)[e] |
|---|---|---|---|---|
| 1 | Breast[a] | 699 | 9 | 2 (241/458) |
| 2 | WDBC[b] | 569 | 30 | 2 (212/357) |
| 3 | WPBC[c] | 198 | 33 | 2 (47/151) |
| 4 | Heart Disease | 303 | 13 | 2 (139/164) |
| 5 | Statlog (Heart) | 270 | 13 | 2 (120/150) |
| 6 | SPECT Heart | 267 | 22 | 2 (55/212) |
| 7 | SPECTF Heart | 267 | 44 | 2 (55/212) |
| 8 | Hepatitis | 155 | 19 | 2 (32/123) |
| 9 | Kidney Disease | 400 | 24 | 2 (250/150) |
| 10 | Thyroid | 215 | 5 | 3 (65[f]/150) |

[a] Breast: Wisconsin Original Breast Cancer.
[b] WDBC: Wisconsin Prognostic Breast Cancer.
[c] WPBC: Wisconsin Prognostic Breast Cancer.
[e] The number of samples in each class is given in the bracket, separated by "/" and presented by "+" for positive (malignant) and "-" for negative (benign) samples.
[f] The two classes with malignant samples are merged as positive samples.

Furthermore, RI and Jaccard measure are also adopted, which are more robust against imbalanced datasets [56]. Especially, the Jaccard measure can better indicate how well the classifier segregates the positive samples, i.e., malignant samples in the medical datasets [57, 58]. RI and Jaccard measure can be calculated using (20a) and (20b) respectively, where $a$ and $b$ denote the number of pairs of examples belonging to the same class that are classified into the same or different predicted classes respectively. $N$ denotes the number of samples.

$$RI = \frac{a+b}{N(N-1)/2} \quad (20a)$$

$$Jaccard = \frac{TP}{TP + FP + FN} \quad (20b)$$

*2) Model Complexity*

The model complexity is evaluated by the number of parameters in the resulting model. In the experiments, zero-order TSK FS is concerned for all the algorithms under comparison. The complexity of the proposed method has been detailed in Section-III-E-1 and the model complexity of the other methods are analyzed as follows.

For TSK-FC, TSK-IQP and TSK-LSSLI, they can be used directly for binary classification. There are $2dK$ parameters for the center $v_i^k$ and the kernel width $\sigma_i^k$ of the Gaussian membership function in the antecedent part, and $K$ parameters for $p^k$ in the consequent part. Therefore, the number of parameters of these three FSs is $(2d+1)K$ for binary classification. Based on the strategy adopted for multi-class classification, there are $m$ zero-order TSK FSs which share the same antecedent part. Hence, the number of parameters for these three models is $(2d+m)K$ when they are used for multi-class classification.

For L2-TSK-FS and TSK-FS-PCA, they can be used directly for regression. When they are used for classification, there will be $m$ zero-order TSK FSs. Hence, the number of parameters of these two models is $m(2d+1)K$.

### D. Classification Performance Evaluation

Fig.1 shows the mean of precision and recall of the six methods under comparison on the ten datasets. It can be seen that the proposed method outperforms the other algorithms in terms of both metrics. The accuracy of the methods is given in Table III. The results show that accuracy of the proposed method is better than, or at least comparable with that of the other methods. Furthermore, the results of F-measure, RI and Jaccard are given in Tables IV~VI. The proposed method also shows superiority even on imbalanced datasets.

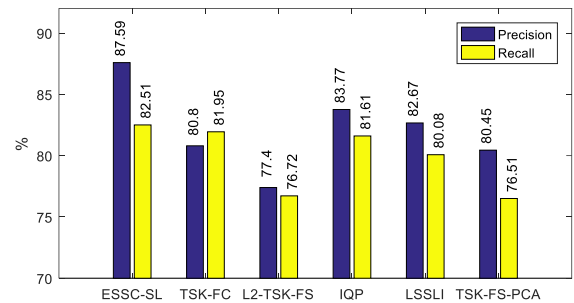

Fig.1. The mean of precision and recall on the ten datasets for six methods.

TABLE III
CLASSIFICATION ACCURACY OF DIFFERENT METHODS ON TEN REAL-WORLD MEDICAL DATASETS

|  | TSK-FC | L2-TSK-FS | TSK-IQP | TSK-LSSLI | TSK-FS-PCA | ESSC-SL-TSK |
|---|---|---|---|---|---|---|
| Breast | 96.30±0084 | 96.44±0079 | 96.50±0112 | 96.58±0088 | 96.50±00102 | **97.14±0113** |
| WDBC | 91.39±0189 | 91.88±0176 | **96.31±0127** | 96.10±0171 | 86.17±0245 | 95.96±0148 |
| WPBC | 75.89±0397 | 76.42±0334 | 77.52±0402 | 77.64±0298 | 76.58±0358 | **81.33±0408** |
| Heart Disease | 78.91±0322 | 82.25±0406 | **82.97±0258** | 82.31±0366 | 77.88±0312 | 82.50±0398 |
| Statlog (Heart) | 79.59±0663 | 83.26±0532 | 81.74±0408 | 80.85±0572 | 79.63±0433 | **83.70±0591** |
| SPECT Heart | 83.64±0432 | 84.35±0516 | 82.73±0373 | 82.25±0601 | 84.35±0524 | **85.39±0530** |
| SPECTF Heart | 79.40±0602 | 79.42±0512 | 80.83±0467 | 80.19±0762 | 79.42±0462 | **80.95±0760** |
| Hepatitis | 81.03±0253 | 79.61±0362 | 81.42±0291 | 81.61±0367 | 80.97±0223 | **85.16±0489** |
| Kidney Disease | 93.10±0127 | 93.43±0253 | 94.08±0176 | 93.50±0089 | 92.75±0180 | **97.75±0205** |
| Thyroid | 95.16±0185 | 95.86±0156 | 96.19±0237 | 96.00±0273 | 96.44±0240 | **97.21±0255** |
| Average | 85.44±0325 | 86.29±0333 | 87.03±0285 | 86.70±0358 | 85.07±0298 | **88.71±0389** |

TABLE IV
CLASSIFICATION F-MEASURE OF DIFFERENT METHODS ON TEN REAL-WORLD MEDICAL DATASETS

|  | TSK-FC | L2-TSK-FS | TSK-IQP | TSK-LSSLI | TSK-FS-PCA | ESSC-SL-TSK |
|---|---|---|---|---|---|---|
| Breast | 96.04±0234 | 96.14±0234 | 96.18±0127 | 96.23±0136 | 96.20±0116 | **96.90±0119** |
| WDBC | 90.67±0126 | 91.40±0242 | **96.11±0143** | 95.98±0180 | 86.51±0143 | 95.74±0151 |
| WPBC | 58.50±0503 | 47.29±0453 | 66.31±0397 | 65.74±0413 | 51.58±0577 | **69.52±0931** |
| Heart Disease | 78.57±0298 | 82.25±0379 | **82.99±0313** | 82.19±0238 | 77.76±0383 | 82.57±0379 |
| Statlog (Heart) | 79.39±0628 | 83.19±0429 | 81.41±0500 | 80.64±0462 | 79.61±0335 | **83.66±0556** |
| SPECT Heart | 73.61±0357 | 72.84±0443 | 70.43±0450 | 68.03±0496 | 72.94±0579 | **78.52±0647** |
| SPECTF Heart | **67.92±0578** | 44.24±0401 | 67.52±0642 | 58.48±0555 | 61.17±0772 | 62.78±0784 |
| Hepatitis | 73.79±0733 | 61.94±0426 | 70.92±0795 | **75.14±0607** | 64.66±0538 | 72.77±0648 |
| Kidney Disease | 92.84±0330 | 93.00±0252 | 93.91±0177 | 93.38±0279 | 92.58±0208 | **97.65±0216** |
| Thyroid | 93.64±0235 | 94.14±0267 | 94.81±0391 | 94.75±0112 | 96.33±0175 | **96.90±0303** |
| Average | 80.50±0402 | 76.64±0353 | 82.06±0394 | 81.04±0348 | 77.93±0384 | **83.70±0473** |

TABLE V
RAND INDEX OF DIFFERENT METHODS ON TEN REAL-WORLD MEDICAL DATASETS

|  | TSK-FC | L2-TSK-FS | TSK-IQP | TSK-LSSLI | TSK-FS-PCA | ESSC-SL-TSK |
|---|---|---|---|---|---|---|
| Breast | 92.94±0251 | 93.10±0271 | 93.50±0357 | 93.48±0360 | 93.51±0456 | **94.72±0317** |
| WDBC | 84.01±0443 | 85.57±0311 | 92.89±0475 | 92.49±0537 | 76.21±0322 | **93.56±0428** |
| WPBC | 63.89±0357 | 64.06±0472 | 66.04±0569 | 65.35±0657 | 64.22±0594 | **69.28±0652** |
| Heart Disease | 66.52±0231 | 70.20±0332 | 71.59±0365 | 70.74±0203 | 65.29±0239 | **71.61±0285** |
| Statlog (Heart) | 67.67±0883 | **72.39±0711** | 70.23±0643 | 68.88±0885 | 66.44±0787 | 71.88±0890 |
| SPECT Heart | 72.88±0988 | 73.50±0859 | 71.54±0820 | 70.24±1033 | **73.55±1144** | 73.29±0834 |
| SPECTF Heart | 67.26±0757 | 67.39±0946 | 69.10±0604 | 68.13±0883 | 67.45±1065 | **69.53±1065** |
| Hepatitis | 68.33±0825 | 67.54±0878 | 70.20±0601 | 70.10±0829 | 69.64±0610 | **79.01±0728** |
| Kidney Disease | 87.26±0325 | 87.52±0481 | 88.84±0234 | 87.84±0340 | 86.56±0313 | **95.61±0395** |
| Thyroid | 93.56±0284 | 93.49±0176 | 93.82±0220 | 93.46±0465 | 95.28±0194 | **96.26±0372** |
| Average | 76.43±0534 | 77.48±0544 | 78.78±0488 | 78.08±0619 | 75.82±0572 | **81.48±0597** |

TABLE VI
JACCARD MEASURE OF DIFFERENT METHODS ON TEN REAL-WORLD MEDICAL DATASETS

|  | TSK-FC | L2-TSK-FS | TSK-IQP | TSK-LSSLI | TSK-FS-PCA | ESSC-SL-TSK |
|---|---|---|---|---|---|---|
| Breast | 86.84±0631 | 90.18±0225 | 90.75±0153 | 89.91±0172 | 90.19±0198 | **92.28±0277** |
| WDBC | 78.33±0168 | 79.63±0293 | 89.67±0132 | 89.85±0385 | 71.52±0254 | **90.41±0417** |
| WPBC | 25.67±0143 | 23.73±0384 | 25.99±0389 | 23.59±0466 | 23.74±0604 | **26.72±0483** |
| Heart Disease | 64.92±0454 | 66.50±0476 | 64.06±0283 | 66.02±0279 | 60.95±0169 | **68.58±0456** |
| Statlog (Heart) | 57.58±0774 | 67.61±0234 | 61.61±0225 | 60.91±0282 | 62.49±0322 | **69.46±0374** |
| SPECT Heart | 41.39±0138 | 40.91±0622 | 29.78±0656 | 25.63±0776 | 31.90±0405 | **43.48±0563** |
| SPECTF Heart | **30.92±0514** | 20.59±0424 | 28.80±0348 | 11.60±0549 | 20.62±0601 | 20.67±0598 |
| Hepatitis | 36.42±0471 | 16.23±0689 | 37.68±0615 | 31.80±0605 | 27.33±0832 | **41.19±0542** |
| Kidney Disease | 90.07± 0248 | 90.59±0311 | 85.85±0791 | 89.20±0114 | 89.62±0309 | **95.36±0172** |
| Thyroid | 94.23± 0417 | 94.61±0060 | 97.47±0086 | 97.48±0147 | **98.85±0533** | 97.14±0639 |
| Average | 60.64±0410 | 59.06±0372 | 61.17±0368 | 58.60±0378 | 57.72±0423 | **64.53±0452** |

TABLE VII
MODEL COMPLEXITY OF THE MODELS OBTAINED BY DIFFERENT METHODS ON TEN REAL-WORLD DATASETS

| --- | TSK-FC | L2-TSK-FS | TSK-IQP | TSK-LSSLI | TSK-FS-PCA | ESSC-SL-TSK |
|---|---|---|---|---|---|---|
| 1 | 570 | 1140 | 570 | 570 | 540 | **174** |
| 2 | 1830 | 3660 | 1830 | 1830 | 1860 | **246** |
| 3 | 2010 | 4020 | 2010 | 2010 | 1980 | **229** |
| 4 | 810 | 1620 | 810 | 810 | 780 | **213** |
| 5 | 810 | 1620 | 810 | 810 | 780 | **197** |
| 6 | 1350 | 2700 | 1350 | 1350 | 1380 | **240** |
| 7 | 2670 | 5340 | 2670 | 2670 | 2700 | **217** |
| 8 | 1170 | 2340 | 1170 | 1170 | 1140 | **176** |
| 9 | 1470 | 2940 | 1470 | 1470 | 1500 | **204** |
| 10 | 1530 | 4410 | 1530 | 1530 | 2250 | **356** |
| --- | **1422** | **2979** | **1422** | **1422** | **1491** | **225.2** |

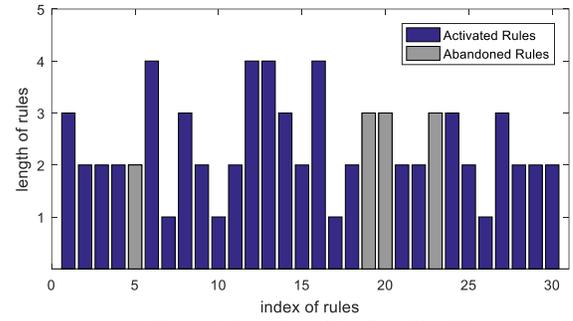

(a) The resulting rules of the first TSK FS

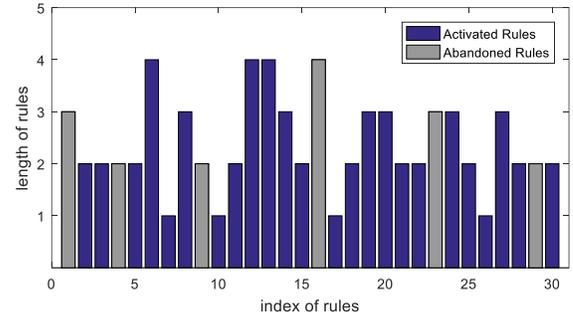

(b) The resulting rules of the second TSK FS

Fig. 2. The resulting rules of the TSK-FS after ESSC and SL.

### E. Interpretability Analysis

The proposed ESSC-SL-CTSK-FS is expected to have better interpretability. Experiments are conducted to verify this advantage from two aspects: quantitative evaluation of model complexity and demonstration of the intuitiveness of the rule-based form.

For the first aspect, the model complexity of the six methods is calculated based on the analysis in Section V-C-2. The results are shown in Table VII. The model complexity of the proposed ESSC-SL-CTSK-FS is calculated based on the model with the best F-measure value in Table IV. It can be concluded from the results that the proposed method outperforms the other algorithms for all the ten datasets. More importantly, the model complexity of the proposed method is significantly lower, where the number of parameters is only about 10% of the other methods.

For the second aspect, the interpretability of the proposed method can be demonstrated by the intuitiveness resulting from its concise rule-based form and human-like fuzzy inference. Taking the Breast dataset as the example. Using the adopted strategy for multi-classification, two TSK FSs are constructed for binary classification. They share the same antecedent parameters while having different consequent parameters. The rules of the resulting models are demonstrated in Fig. 2. The *x*-axis denotes the indices of all the rules in the rule base and there are 30 rules at the beginning. The *y*-axis denotes the length of rules which is decided by the number of selected features for each rule. It can be seen from Fig. 2(a) and 2(b) that only 1 to 4 features are remained after applying ESSC, i.e., the length of the rules is reduced. Given the capacity of human cognition is limited, the proposed method produces more interpretable results. Besides, some redundant rules are abandoned after applying SL. For the first FS in Fig. 2(a), the indices of the abandoned rules are 5, 19, 20 and 23, whereas for the second FS in Fig. 2(b), the indices of the abandoned rules are 1, 4, 9, 16, 23 and 29. Hence, as demonstrated by Fig. 2, the proposed ESSC-SL-CTSK-FS enables the construction of concise FS and improves interpretability.

To illustrate the rules of the resulting model more specifically, the linguistic descriptions of the rules are given as follows. Firstly, Fig. 3 shows the activation of the selected feature subsets. The *x*-axis and *y*-axis are the indices of rules and the indices of features respectively. For the Breast dataset, there are 9 features. Each block represents a feature. The colors of the blocks indicate whether a feature is selected or abandoned (i.e. gray). To demonstrate the linguistic descriptions of the rules, the centers of the membership function in (5) are divided into five intervals, each represented with a distinct color as shown in the figure. Since the values of the datasets is normalized into [0,1], the intervals of the centers are also divided within the same range, i.e., the five intervals are [0,0.2], [0.2,0.4], [0.4,0.6], [0.6,0.8] and [0.8,1], which correspond to the five colors and the five vague semantics: *Low*, *Lower*, *Medium*, *Higher* and *High*. It can be seen from Fig. 3 that the activation of the features in different rules is sparse and scattered.

For the first TSK FS in Fig. 3 that is generated using the Breast dataset, the fuzzy rule base can be described with Table VIII which demonstrates that the rules generated with the proposed method are more interpretable

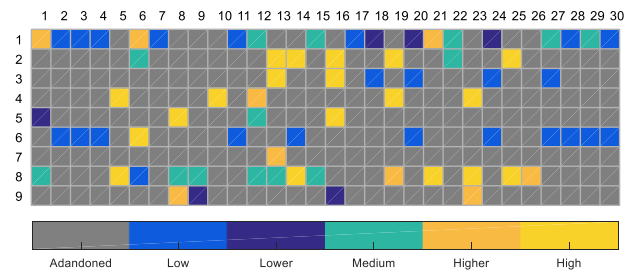

Fig. 3 The degree of activation of the features in the rule base

TABLE VIII
THE RULE BASE GENERATED ON BREAST DATASET

| The Rule Base of the Breast Dataset |
| --- |
| **Rule 1:**<br>IF: the *first* feature is *Higher*, and<br>the *fifth* feature is *Lower*, and<br>the *eighth* feature is *Medium*.<br>Then: the output of the rule is -0.0112.<br><br>**Rule 2:**<br>IF: the *first* feature is *Low*, and<br>the *sixth* feature is *Low*<br>Then: the output of the rule is 0.3470.<br><br>…… (These rules can be obtained according to Fig.3 and the rules 5, 19, 20 and 23 are abandoned according to Fig.2 (a)).<br><br>**Rule 29:**<br>IF: the *first* feature is *Medium*, and<br>the *sixth* feature is *Low*<br>Then: the output of the rule is 1.4885.<br><br>**Rule 30:**<br>IF: the *first* feature is *Low*, and<br>the *sixth* feature is *Low*<br>Then: the output of the rule is 3.9374. |

### F. Effectiveness of the ESSC and Sparse Learning

In this section, the model complexity of the proposed ESSC-SL-CTSK-FS is analyzed under different settings of ESSC and SL. All the experiments conducted on the Breast dataset with $h=10$, $\varepsilon=0.01$ and $\eta=0.01$.

The most important parameter of ESSC is the threshold $\beta$ used to select the features. Given the weight vector $\mathbf{w}_k = [w_{k1},\cdots,w_{kd}]$ obtained that corresponds to the contribution of all the features to the $k$th rule, the threshold $\beta$ is usually set to a value between [0,1] and the features with the weights $w_{kl} > \beta$ are selected. In Sections D and E, the experiments are conducted with $\beta \geq 0.1$. Here, $\beta$ is within [0,1] for a more extensive evaluation of its influence.

The regularization parameter $\lambda$ is the key to control the sparsity of the consequent parameters of fuzzy systems, i.e., the number of the rules in the rule base. The parameter is optimally set by the search grid $\{0.1,0.2,\cdots,0.8,0.9\}$ in Sections D and E. In this section, the sparsity of the rules is studied more closely with various parameter settings.

In the following three subsections, the analysis of model complexity of FS constructed based on ESSC and SL is conducted from three aspects: FS constructed by the ESSC technique only, FS constructed by the SL technique only, and the FS constructed by ESSC and SL simultaneously.

#### 1) Effectiveness of ESSC

By keeping all the other parameters fixed and setting the sparsity regularization parameter $\lambda$ to 0, the model complexity and the classification performance are analyzed with different values of the weight threshold of ESSC. The threshold is set as $\{0,0.05,0.1,0.15,0.2,0.25,0.3\}$. Fig. 4 shows the activation of the features in different rules under different thresholds. The gray blocks denote the abandoned features and the blue blocks denote the activated features. Note that the format of the subfigures in Fig. 4 is the same as that of Fig. 3; they are rotated and put together for compact display and easy comparison. It can be seen that the sparsity of the activated features increases with the threshold.

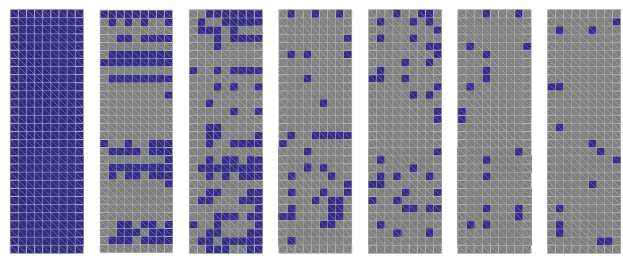

(a): 0  (b): 0.05  (c): 0.1  (d): 0.15  (e): 0.2  (f): 0.25  (g): 0.3

Fig. 4. The activation of features by ESSC under different weight threshold

Fig. 5 illustrates the number of parameters and the corresponding F-measure of the obtained FSs using the different values of the weight threshold. The figure shows that the number of the parameters of the resulting model decreases with increasing threshold, which is consistent with the results in Fig. 4. Besides, the value of F-measure decreases with decreasing number of parameters. It can be concluded that the weight threshold can reduce the number of parameters effectively, demonstrating the ability of ESSC in feature selection and concise FS construction.

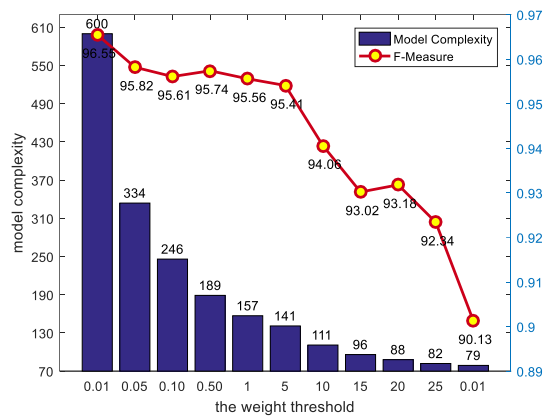

Fig.5. The F-measure and number of parameters under different weight thresholds

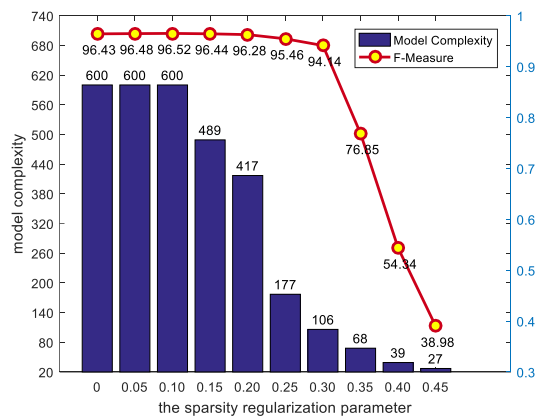

Fig.6. The F-measure and number of parameters under different sparsity regularization parameter

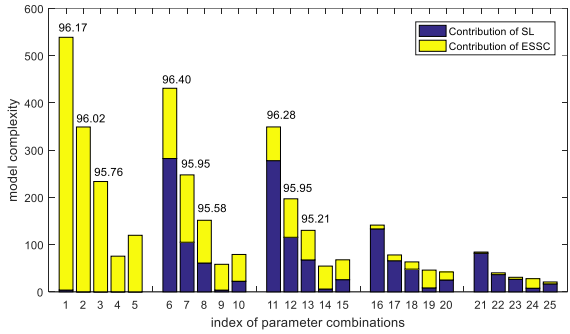

Fig.7. The contributions of ESSC and SL with different parameter combinations

*2) Effectiveness of SL*

To analyze the effectiveness of SL, the sparsity regularization parameter $\lambda$ is varied while the other parameters fixed. The weight threshold is set to 0, which means that all the features in the rule base are used. Fig. 6 shows the F-measure and the number of parameters under different sparsity regularization parameters. It can be seen that the F-measure almost remains the same or becomes even higher when the sparsity regularization parameter is less than 0.25, whereas the number of parameters is reduced considerably. The above analysis shows that SL can effectively reduce model complexity while maintaining the classification performance.

*3) Effectiveness of combining ESSC and SL*

As discussed in Section I, classification performance usually increases with model complexity. The proposed method aims to reduce the model complexity while keeping the classification performance. The key is to integrate ESSC and SL with an ideal combination of their parameters. An experiment is conducted here to study the effect of different combinations of the weight threshold parameter of ESSC and the sparsity regularization parameter of SL. The weight threshold is taken from the search grid {0.01,0.05,0.1,0.5,1} of 5 values and varied within each group. Similarly, the sparsity regularization parameter is taken from the search grid {0.1,0.5,1,5,10} of 5 values and varied between the groups. That is, a search grid of 25 combinations of the two parameters is used.

Fig. 7 shows the model complexity and F-measure obtained with different parameter combinations. The figure only displays the results of combinations that achieve 95% F-measure or above. The stacked bars are used to display the contributions of ESSC and SL in reducing the model complexity of the resulting FSs. The contributions are obtained by calculating the proportion of the reduced number of parameters to the total number of parameters for ESSC and SL respectively. It can be seen in the first group of stacked bars on the left, with indices 1 to 5, that the reduction in model complexity is mostly attributed to ESSC, while SL almost has no effect on fuzzy rule reduction since the sparsity regularization parameter is set to 0.1. In the fourth and fifth groups of stacked bars on the right, SL is dominated in making contribution to model complexity reduction, but the classification performance and model complexity are both low because there are only few rules in the rule base. The second and the third groups of staked bars produce more satisfactory results with a more balanced contribution of ESSC and SL, where low model complexity and high classification performance is achieved. The experiment shows that optimal parameter combinations are a key factor. In practice, the best parameter settings can be decided according to the acceptable degree of trade-off between the classification performance and model complexity.

TABLE IX
FRIEDMAN TEST ON THE RESULTS OF F-MEASURE

| Algorithm | Ranking | *p*-value | Hypothesis |
|---|---|---|---|
| TSK-FC | 4.55 | | |
| L2-TSK-FS | 3.8 | | |
| TSK-IQP | 3.35 | 0.002673 | Reject |
| TSK-LSSLI | 3.4 | | |
| TSK-FS-PCA | 4.65 | | |
| ESSC-SL-TSK | 1.25 | | |

TABLE X
FRIEDMAN TEST ON THE RESULTS OF MODEL COMPLEXITY

| Algorithm | Ranking | *p*-value | Hypothesis |
|---|---|---|---|
| TSK-FC | 3.5 | | |
| L2-TSK-FS | 6 | | |
| TSK-IQP | 3.5 | 0.000001 | Reject |
| TSK-LSSLI | 3.5 | | |
| TSK-FS-PCA | 3.5 | | |
| ESSC-SL-TSK | 1 | | |

The results show that the proposed method is a flexible TSK FS construction method. The importance of ESSC and SL can be readily adjusted by selecting different parameter settings according to the requirements of the application scenarios. Although there exist some other methods that can also be used to reduce the model complexity of TSK FS, they lack the flexibility of the proposed method. For example, the methods usually attempt to reduce model complexity by employing the strategy of grid search for the number of rules, given the upper limit of the number of rules in the rule base. However, the strategy is deficient in that it can only reduce the model complexity from the aspect of rule reduction and cannot deal with high-dimensional data, where optimal feature selection is necessary for controlling the length of rules and achieving good interpretability. Besides, the strategy is indeed an exhaustive search that is computationally intensive and cannot select the rules adaptively in a way like the proposed method with SL. Other methods such as random feature selection and subspace clustering as discussed in Section II-B have also been used to reduce model complexity, but all of these existing methods do not have the ability to construct the concise and compact TSK FS from the aspects of feature selection and rule reduction simultaneously.

*G. Statistical Analysis*

To further evaluate the effectiveness of the proposed method, statistical tests are conducted to analyze the significance of the experimental results. The non-parametric Friedman test [59] is used to determine whether the results obtained by the six methods under comparison are significantly different. In the Friedman test, the value of $\alpha$ is set as 0.05 such that if the $p$-value is less than $\alpha$, the null hypothesis that the performance of all the algorithms is the same is rejected. The post-hoc test is then used to further determine whether the performance of the best method, as identified by the Friedman test, is significantly different from that of the other methods.

TABLE XI
POST-HOC TEST ON THE RESULTS OF F-MEASURE

| $i$ | Algorithm | $z = (R_0 - R_i)/SE$ | $p$ | Holm= $\alpha/i$ | Hypothesis |
|---|---|---|---|---|---|
| 5 | L2-TSK-FS | 3.34664 | 0.000818 | 0.01 | Reject |
| 4 | TSK-FS-PCA | 3. 34664 | 0. 000818 | 0.0125 | Reject |
| 3 | TSK-FC | 2.988072 | 0.002807 | 0.016667 | Not Reject |
| 2 | TSK-LSSLI | 2.031889 | 0.042165 | 0.025 | Not Reject |
| 1 | TSK-IQP | 1.195229 | 0.231998 | 0.05 | Not Reject |

TABLE XII
POST-HOC TEST ON THE RESULTS OF MODEL COMPLEXITY

| $i$ | Algorithm | $z = (R_0 - R_i)/SE$ | $p$ | Holm= $\alpha/i$ | Hypothesis |
|---|---|---|---|---|---|
| 5 | L2-TSK-FS | 5.976143 | 0 | 0.01 | Reject |
| 4 | TSK-FS-PCA | 2.988072 | 0.002807 | 0.0125 | Reject |
| 3 | TSK-FC | 2.988072 | 0.002807 | 0.016667 | Reject |
| 2 | TSK-LSSLI | 2.988072 | 0.002807 | 0.025 | Reject |
| 1 | TSK-IQP | 2.988072 | 0.002807 | 0.05 | Reject |

Tables IX and X show the results of Friedman test on F-measure and model complexity, which indicate that the performance of the six methods is significantly different and that the proposed ESSC-SL-CTSK-FS is ranked first, superior to the other methods.

Based on the results of Friedman test, the post-hoc test is conducted to compare the best method, i.e., ESSC-SL-CTSK-FS, with each of the other methods regarding their F-measure performance and model complexity respectively. The results of the post-hoc test are shown in Tables XI and XII. From the aspect of F-measure, the results in Table XI show that the proposed ESSC-SL-CTSK-FS is significantly superior to L2-TSK-FS and TSK-FS-PCA, but not so for the other three methods. Nevertheless, it can be seen from Table IV that ESSC-SL-CTSK-FS still outperforms these three methods to some extent. From the aspect of model complexity, Table XII shows that ESSC-SL-CTSK-FS is significantly better than all the other algorithms.

The results of the statistical analysis presented above show that the purpose of the proposed method is met, i.e., to construct concise and interpretable FS while keeping the classification performance competitive with, or even better than conventional fuzzy models.

## VI. CONCLUSIONS

Interpretability of decision model is very important in many practical applications, such as medical diagnosis. To meet the requirement, this paper investigates the development of highly interpretable intelligent models based on concise zero-order TSK FS. Two techniques are used to improve the interpretability of zero-order TSK FS. First, ESSC is adopted to partition the input space of the training dataset. Elastic fuzzy rules can then be generated where each rule only contains a few important features and different rules are constructed with different feature subsets. Thus, the concise fuzzy rules not only remove the noisy features but also possess human-like inference mechanism that consider different views from different experts on the same task. Second, SL is adopted to remove redundant rules by solving the sparse solution to the consequent parameters of the fuzzy rules. With the two techniques, the ESSC-SL-CTSK-FS method is proposed to construct concise and highly interpretable zero-order TSK FS, which have demonstrated promising performance with extensive experiments conducted on various medical datasets.

Further research of the project includes the investigation of using other FS models, e.g. first order TSK FS and Mamdani-Larsen-type FS, to improve high-dimensional data-driven fuzzy modeling in terms of interpretability and conciseness. Another interesting work is to investigate compact subspace extraction methods for ESSC-SL-CTSK-FS when the input features in different rules are almost equally important.

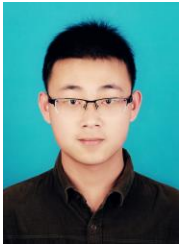

**Peng Xu** received the B.S. degree in computer science from Jiangnan University, Wuxi, China, in 2017. He is currently a master student in the School of Digital Media, Jiangnan University. His research interests include computational intelligence, machine learning, interpretable artificial intelligence, fuzzy modeling and their applications.

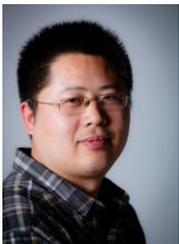

**Zhaohong Deng** (M'2012-SM'2014) received the B.S. degree in physics from Fuyang Normal College, Fuyang, China, in 2002, and the Ph.D. degree in information technology and engineering from Jiangnan University, Wuxi, China, in 2008. He is currently a professor in the School of Digital Media, Jiangnan University. He has visited University of California-Davis and the Hong Kong Polytechnic University for more than two years. His current research interests include uncertainty modeling, neuro-fuzzy systems, pattern recognition, and their applications. He is the author or coauthor of more than 100 research papers in international/national journals. He has served as an associate editor or guest editor of several international Journals, such as IEEE Transactions on Emerging Topics in Computational Intelligence, Neurocomputing and PLOS ONE.

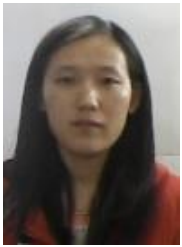

**Chen Cui** received the M.S. degree in software engineering from the School of Digital Media, Jiangnan University, Wuxi, China, in 2018. Her research interests include intelligent computation, fuzzy modeling and their applications.

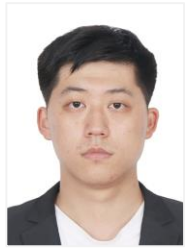

**Te Zhang** received the M.S. degree in software engineering from the School of Digital Media, Jiangnan University, Wuxi, China, in 2018. He is currently a Research Assistant with the Department of Computer Science and Engineering, Southern University of Science and Technology, Shenzhen, China. His current research interests include machine learning, fuzzy systems, data mining, intelligent computation.

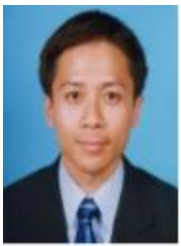

**Kup-Sze Choi** (M'97) received the Ph.D. degree in computer science and engineering from the Chinese University of Hong Kong. He is currently an Associate Professor at the School of Nursing, Hong Kong Polytechnic University and the Director of the Centre for Smart Health. His research interests include virtual reality and artificial intelligence, and their applications in medicine and healthcare.

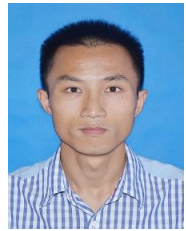

**Suhang Gu** is a Ph.D. candidate at the School of Digital Media, Jiangnan University. He received his M.S. degree in the School of Information Science and Engineering of Changzhou University in 2015. He is a Lecturer with the School of Information Engineering and Technology, Changzhou Vocational Institute of Light Industry, Changzhou, China. He has published some papers in national authoritative journals. His main research interests include pattern recognition and data mining.

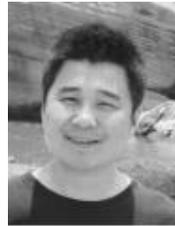

**Jun Wang** (M'14) received his Ph.D. degree in pattern recognition and intelligence systems from the School of Computer Science and Technology in Nanjing University of Science and Technology, Nanjing (NUST), China, in 2011. He has been a Research Assistant in the Department of Computing, Hong Kong Polytechnic University, China, and a postdoc research fellow in the Department of Radiology and BRIC, School of Medicine, University of North Carolina at Chapel Hill, USA, respectively. He is currently an Associate Professor with the School of Digital Media, Jiangnan University, China. He has published more than 50 articles in international/national journals. His research interests include machine learning, fuzzy systems and medical image classification.

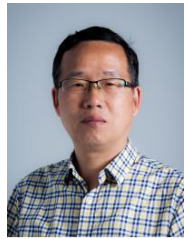

**Shitong Wang** received the M.S. degree in Computer Science from Nanjing University of Aeronautics and Astronautics, China, in 1987. His research interests include artificial intelligence, neuro-fuzzy systems, pattern recognition, and image processing. He has published about 100 papers in international/national journals and has authored seven books.